\documentclass[conference, 10pt]{IEEEtran}
\usepackage{array}
\newcolumntype{x}[1]{>{\centering\arraybackslash\hspace{0pt}}p{#1}}
\usepackage[numbers,sort&compress]{natbib}
\usepackage[colorlinks = true,
            linkcolor = black,
            urlcolor  = blue,
            citecolor = magenta,
            anchorcolor = black,
            bookmarks = true]{hyperref}

\usepackage{diagbox}
\usepackage{graphicx}
\graphicspath{{figures/}}
\usepackage{hhline}
\usepackage{bbm}
\let\labelindent\relax
\usepackage{enumitem}
\usepackage{amsmath}
\usepackage{amssymb}
\usepackage{lipsum}
\usepackage{xcolor}

\newcommand{\figtop}{{\em (Top)}}
\newcommand{\figbottom}{{\em (Bottom)}}

\def\eqref#1{equation~\ref{#1}}

\def\1{\bm{1}}

\DeclareMathAlphabet{\mathsfit}{\encodingdefault}{\sfdefault}{m}{sl}
\SetMathAlphabet{\mathsfit}{bold}{\encodingdefault}{\sfdefault}{bx}{n}

\def\gB{{\mathcal{B}}}

\def\gM{{\mathcal{M}}}

\def\gP{{\mathcal{P}}}

\def\gT{{\mathcal{T}}}

\def\sD{{\mathbb{D}}}

\usepackage{algorithm}
\usepackage{algpseudocode}
\usepackage{booktabs}
\usepackage[size=footnotesize, labelfont=bf]{caption}

\newcommand{\sysName}{ViNG~}
\newcommand{\sysNamenosp}{ViNG}
\begin{document}

\title{\huge ViNG: Learning Open-World Navigation with Visual Goals}

\DeclareRobustCommand*{\IEEEauthorrefmark}[1]{%
  \raisebox{0pt}[0pt][0pt]{\textsuperscript{\footnotesize #1}}%
}

\author{\IEEEauthorblockN{Dhruv Shah\IEEEauthorrefmark{1}, Benjamin Eysenbach\IEEEauthorrefmark{2}, Gregory Kahn\IEEEauthorrefmark{1}, Nicholas Rhinehart\IEEEauthorrefmark{1}, Sergey Levine\IEEEauthorrefmark{1}}
 \IEEEauthorblockA{%
    \IEEEauthorrefmark{1}UC Berkeley, 
    \IEEEauthorrefmark{2}Carnegie Mellon University
  }}

\maketitle
\begin{abstract}

We propose a learning-based navigation system for reaching visually indicated goals and demonstrate this system on a real mobile robot platform.
Learning provides an appealing alternative to conventional methods for robotic navigation: instead of reasoning about environments in terms of geometry and maps, learning can enable a robot to learn about navigational affordances, understand what types of obstacles are traversable (e.g., tall grass) or not (e.g., walls), and generalize over patterns in the environment. However, unlike conventional planning algorithms, it is harder to change the goal for a learned policy during deployment. We propose a method for learning to navigate towards a goal image of the desired destination.
By combining a learned policy with a topological graph
constructed out of previously observed data, our system can determine how to reach this visually indicated goal even in the presence of variable appearance and lighting.
Three key insights, waypoint proposal, graph pruning and negative mining, enable our method to learn to navigate in real-world environments using only offline data, a setting where prior methods struggle.
We instantiate our method on a real outdoor ground robot and show that our system, which we call \sysNamenosp, outperforms previously-proposed methods for goal-conditioned reinforcement learning, including other methods that incorporate reinforcement learning and search. We also study how \sysName generalizes to unseen environments and evaluate its ability to adapt to such an environment with growing experience. Finally, we demonstrate \sysName on a number of real-world applications, such as last-mile delivery and warehouse inspection. We encourage the reader to visit the project website for videos of our experiments and demonstrations\footnote{Project website: \texttt{\href{https://sites.google.com/view/ving-robot}{sites.google.com/view/ving-robot}}}.
\end{abstract}

\IEEEpeerreviewmaketitle

\section{Introduction}
\label{sec:intro}

Visual navigation in complex environments poses several challenges: (i) difficulty in faithfully modeling the complex dynamics and nuanced environmental interactions; (ii) reacting to high-dimensional observations; (iii) cost and safety constraints on collecting data, requiring learning from previously collected (i.e., ``offline'') experience; and (iv) generalizing effecticely across different settings and environments.
Planning algorithms achieve many of these desiderata, but their efficacy depends on having the right representation of the task; it remains unclear how to apply many planning algorithms to tasks with image-based observations.
On the other hand, humans seemingly have little difficulty navigating complex environments from first-person observations, without GPS or maps, if they have seen the environment before. Humans and animals are known to use ``mental maps'' that rely on landmarks and other cues~\cite{okeefe1978map, gillner1998map, foo2005map}, and rely heavily on learning.
Further, in the absence of spatial positional information (e.g., GPS or maps), specification of a navigational goal itself becomes challenging, since locational goals require the robot to be able to compare its location to the target.

\begin{figure}
    \centering
    \includegraphics[width=0.9\columnwidth]{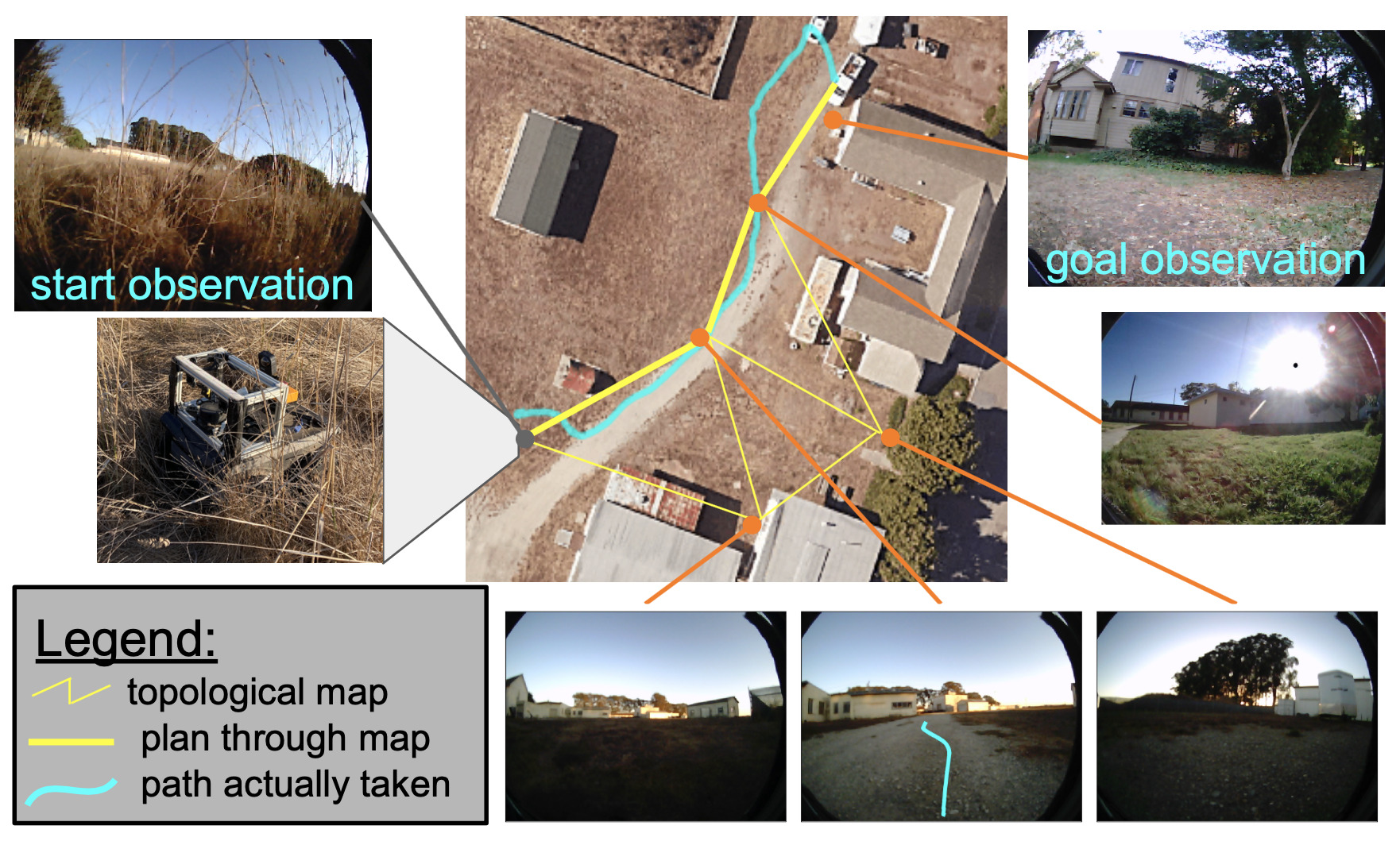}
    \caption{\sysName builds and plans over a \textit{learned} topological graph consisting of previously seen egocentric images, and uses a learned controller to execute the path to a visually indicated goal.
    Unlike prior work, our method uses purely offline experience and does not require a simulator or online data collection.
    Note that the graph constructed by our algorithm is not geometric    and nodes are \emph{not} associated with coordinates in the world, but only with \emph{image observations} -- the top-down satellite image is provided only for visualization and is not available to our method.}
    \label{fig:teaser}
\end{figure}

In this paper, we study learning-based methods for navigation that can similarly utilize graph-structured ``mental maps'' that are non-geometric in nature, and can enable a robot to navigate in the real-world. We use a natural and intuitive mechanism for specifying goals -- where the user provides the robot with a picture of the desired destination. Inspired by humans navigating toward previously seen landmarks, our goal is to enable the robot to navigate to a visually indicated goal. Crucially, such a goal specification scheme does not presume any prior geometric knowledge of the scene, while still providing enough information for the robot to perform the task. Fig.~\ref{fig:teaser} shows an example of such a task.

Towards satisfying these requirements, we present a fully autonomous, self-supervised mobile robot platform for visual goal-reaching in outdoor, unstructured environments which we call \sysName -- visual navigation with goals.
Our approach combines the strengths of dynamical distance learning and graph search. We first learn a function that predicts the \emph{dynamical} distance between pairs of observations, estimating how many time steps are needed to transition between them. We then use this learned dynamical distance to embed past observations into a topological graph, and plan over this graph.
This process makes no geometric assumptions about the environment: reachability is determined entirely by learning from data.
Unlike pure planning-based approaches, our method scales to high-dimensional observations and hard-to-model dynamics,
and does not assume access to any ground-truth spatial information.
Unlike pure learning-based approaches, our method effectively learns from offline experience and reasons over long horizons. Unlike prior methods that combine planning and learning,
\sysName learns from offline, real-world data, and does not require a simulator or online data collection.

The primary contribution of this work is a self-supervised robotic system,
\sysNamenosp, that can efficiently learn goal-directed navigation behaviors in open-world environments without access to spatial maps from an offline pool of data, including randomly collected trajectories.
Three key ideas, waypoint proposal, graph pruning and negative mining, differentiate our method from prior work and are critical to the success of our method in this offline setting.
\sysName can learn to navigate to an arbitrary user-specified visual goal in a variety of open-world settings, including urban, grassy, and rocky terrain, learning only from offline data. Our experiments show that \sysName learns goal-conditioned behaviors that can effectively plan over long horizons. We show that \sysName outperforms several competitive offline RL and geometric baselines.
Further, we show that the learned behaviors transfer to novel environments using as little as 20 minutes of data from the environment and that \sysName can adapt in such novel environments as it gathers more data, resulting in an autonomous, self-improving system. Lastly, we demonstrate two real-world applications enabled by \sysName in dense, urban neighborhoods -- last-mile delivery of food or mail, and autonomous inspection of warehouses.

\section{Related Work}
\label{sec:related}

Prior work has studied vision-based mobile robot navigation in many real-world settings, including indoor and outdoor navigation~\cite{Rosen1968, thorpe1988vision, borenstein1989real}, autonomous driving~\cite{thrun2006stanley, urmson2008autonomous}, and navigation in extra-terrestrial and underwater environments~\cite{krotkov1995mapping, dalgleish2004vision}.
The combination of mapping~\cite{thrun2006probalistic} and path planning~\cite{lavalle2006planning} has been a cornerstone for a number of effective systems~\cite{davison1998mobile, Sim2006, furgale2010visual} and underlies several state-of-the-art navigation systems~\cite{bansal2018chauffeurnet, ackerman2018skydio}.
Many prior methods make restrictive assumptions, such as access to LIDAR or other structured sensor information and accurate localization, which can limit their suitability for deployment in unstructured environments~\cite{subT}. Further, prior work often assumes that geometric traversability is faithfully indicated through observations and not misled by (say) non-obstacles such as tall grass~\cite{fuentes2015visual}.
Learning-based systems lift some of these assumptions
and can use learned models to perform perception~\cite{chen2015deepdriving, Wang2019}, planning~\cite{kahn2018gcg, kumar2018visual, Hartikainen2020}, or both~\cite{levine2016end}. In practice, learning temporally extended long-horizon skills with either reinforcement learning (RL) or imitation learning (IL) remains difficult~\cite{ross2011reduction, dulac2019challenges}.

Recent methods address limitations of the above approaches by combining planning and learning~\cite{eysenbach2019sorb, savinov2018sptm, chaplot2020ans,chiang2019learning, faust2018prmrl, meng2020scaling}. These methods use learning (i.e., approximate dynamic programming) to solve short-horizon tasks and plan (i.e., use exact dynamic programming) over non-metric topological graphs~\cite{Meng1993, Meng1995} to reason over longer horizons. This general approach simultaneously avoids the need for (1) high-fidelity map building and (2) learning temporally-extended behaviors from scratch. However, prior instantiations of this recipe make assumptions that limit their applicability to real-world settings: assuming access to an exact simulation replica of the environment~\cite{francis2020longrange, meng2020scaling}, assuming simplified action spaces~\cite{eysenbach2019sorb, savinov2018sptm, chaplot2020ans}, or requiring online data collection~\cite{eysenbach2019sorb, chaplot2020ans}.
Our experiments in Section~\ref{sec:experiemnts} demonstrate that prior methods fail when they are not allowed to collect new experience in a simulator or the real-world.

Our method, \sysNamenosp, builds on these prior approaches by adding two key ideas: graph pruning and negative sampling. These additional ingredients allow \sysName to lift assumptions made by prior methods: it does not assume access to a simulator, and does not require interactive access to an environment; it is trained using offline, real-world data; and it operates directly on high-dimensional images and predicts continuous actions for the robot. 
To the best of our knowledge, \sysName is the first system demonstrated on a real-world ground robot that can learn from offline data to reach visually indicated navigational goals over long time horizons without simulated training or hand-designed localization and mapping systems.

\section{Problem Statement and System Overview}
\label{sec:setup}

\begin{figure}
    \centering %
    \includegraphics[width=0.9\columnwidth]{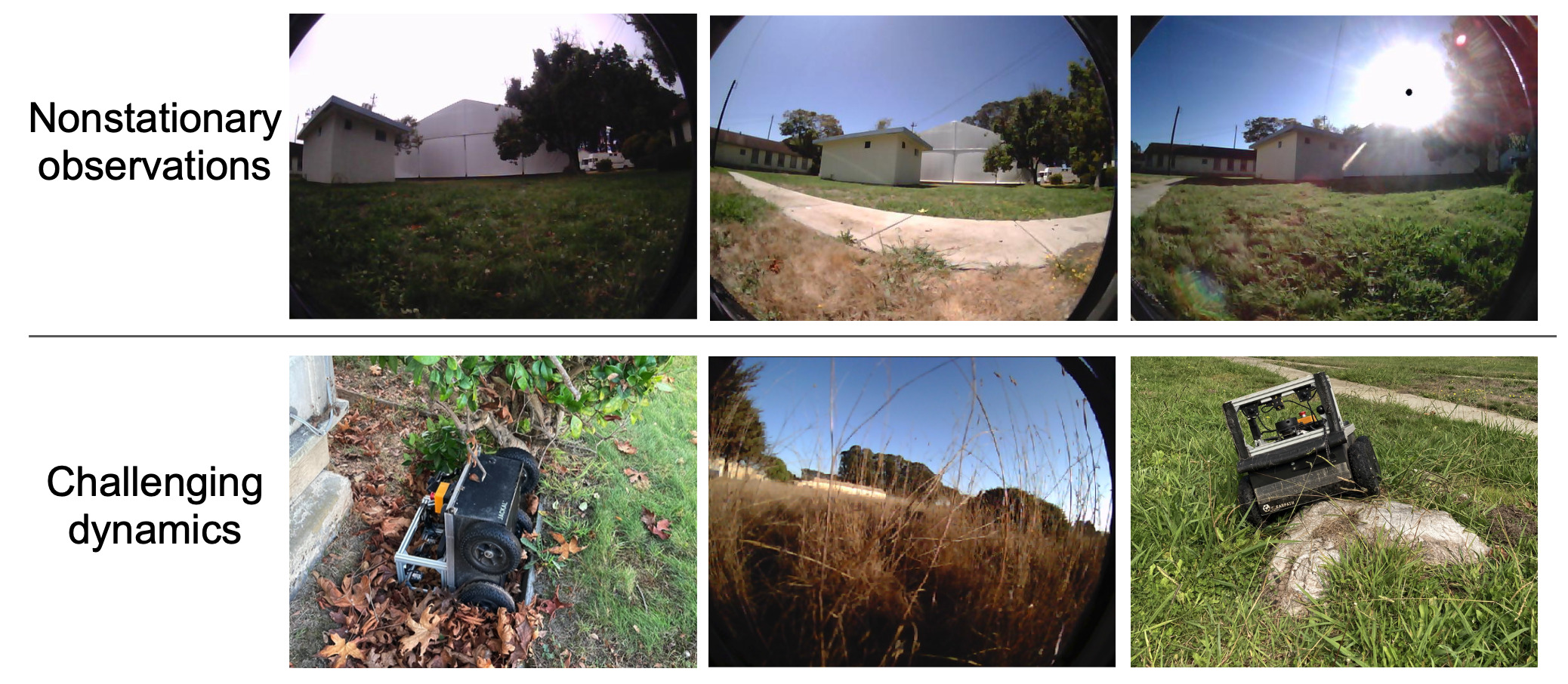}
    \caption{\textbf{Challenges with Real-World Navigation}: \figtop \; Three observations taken from \emph{exactly the same position} at different times of day exhibit large differences. \figbottom \; While tall grass and inclined rocks are traversable, a hole filled with dry leaves is not. These examples highlight the challenges with geometric reasoning about traversability.}
    \label{fig:realworld}
    \vspace*{-1.5em}
\end{figure}

We consider the problem of goal-directed visual navigation: a robot is tasked with navigating to a goal location $G$ given an image observation $o_G$ taken at $G$. In addition to navigating to the goal, the robot also needs to recognize \emph{when} it has reached the goal, signaling that the task has been completed.
The robot does not have a spatial map of the environment, but we assume that it has access to a small number of trajectories that it has collected previously. %
This data will be used to construct a graph over the environment using a learned distance and reachability function.
We make no assumptions on the nature of the trajectories: they may be obtained by human teleoperation, self-exploration, or a result of a random walk. Each trajectory is a dense sequence of observations $o_1, o_2, \ldots, o_n$ recorded by its on-board camera. Since the robot only observes the world from a single on-board camera and does not run any state estimation, our system operates in a partially observed setting. Our system commands continuous linear and angular velocities.

\subsection{Mobile Robot Platform}
\label{sec:jackal_system}
We implement \sysName on a Clearpath Jackal UGV platform -- a small, fast, weatherproof outdoor ground robot ideal for navigating in both urban and off-road environments (see Fig.~\ref{fig:teaser} and \ref{fig:realworld}). The default sensor suite consists of a 6-DoF IMU, a GPS unit for approximate global position estimates,
and wheel encoders to estimate local odometry. In addition, we added a forward-facing $170^\circ$ field-of-view camera and an RPLIDAR 2D laser scanner. Inside the Jackal is an NVIDIA Jetson TX2 computer.
While the robot carries a GPS and laser scanner, we use these sensors solely as a safety mechanism during data collection. Our method solely operates using images taken from the onboard camera.

\subsection{Data Collection \& Labeling}
\label{sec:data_collection}

\sysName can learn navigational behaviors from previously-collected, off-policy data -- a desideratum of real-world robots.
To demonstrate this capability, we run our core experiments using data exclusively from prior work~\cite{kahn2020badgr}; we also collect a limited amount of additional data for our environment generalization experiments using the same self-supervised data collection strategy. The prior data was collected more than 10 months prior to the experiments in this paper (see Fig.~\ref{fig:realworld} \textit{(top)}), and exhibits significant differences in appearance, lighting, time of year, and time of day as compared to the evaluation setting. 
This underscores the ability of \sysName to utilize offline data from diverse sources.

\section{Visual Navigation with Goals}
\label{sec:method}

We approach the problem of visual goal-conditioned navigation by combining non-metric maps and %
learned, image-based, goal-conditioned policies.
We describe our method in two stages: (i) training two learned functions and (ii) deploying the system, which entails using the learned functions together with past experience to execute goal-directed behavior.

During \emph{training}, we use previously collected experience to learn an environment-independent traversability function $\gT$, as well as a relative pose predictor, $\gP$.
During \emph{deployment},
the robot builds a topological graph of its environment: a directed graph with vertices as observations and edges encoding traversability and proximity. At each time step $t$, the robot localizes its current and goal observations ($o_t$, $o_G$) in the graph and follows the best path to $G$, as determined by a graph search algorithm that outputs the next waypoint for the controller.
To close the loop, we need a goal-conditioned controller that takes the current and goal observations, and outputs an action $a$. %
The controller progressively follows the path directed by the planner until it reaches $G$.

While the general recipe of \sysName is similar to prior work~\citep{savinov2018sptm, meng2020scaling, eysenbach2019sorb}, our experiments demonstrate that two key technical insights contribute to significantly improved performance in the real-world setting: \emph{graph pruning} (Sec.~\ref{sec:pruning}) and \emph{negative mining} (Sec.~\ref{sec:negatives}). 
Our comparisons to prior methods in Section~\ref{sec:experiemnts} and ablation studies in Section~\ref{sec:ablations} demonstrate these novel improvements enable \sysName to learn goal-conditioned policies entirely from offline data, avoiding the need for simulators and online sampling, while prior methods struggle to attain good performance, particularly for long-horizon goals.

\begin{algorithm}[t]
\caption{Training \sysName}\label{alg:training}
\begin{algorithmic}[1]
\State  {\footnotesize \textbf{Input} transitions $\{\tau^{(k)} = (o_1^{(k)}, a_1^{(k)}, o_2^{(k)}, a_2^{(k)}, \cdots)\}_{k=1, \cdots}$ }
\State $\mathbbm{D}_+ \gets \{(o_i^{(k)}, o_j^{(k)}, d = \min(j - i, d_\text{max}))\}_{i \le j, k = 1, \cdots}$
\State $\mathbbm{D}_- \gets \{(o_i^{(k)}, o_j^{(\ell)}, d = d_\text{max})\}_{i, j, k \neq \ell}$
\State Initialize $\gT(o_i, o_j)$ and $\gP(o_i, o_j)$
\While{not converged}
\State $\gB_+ \sim \mathbbm{D}_+, \gB_- \sim \mathbbm{D}_-$ \Comment{Sample batch.}
\State $\gT \gets \text{UpdateDistanceFn}(\gT; \gB_+ \cup \gB_-)$
\State get relative pose: $\mathbbm{D}_+ \gets \{((o_i^{(k)}, o_j^{(k)}, d_{ij}, p_{ij}\}$ \label{line:rel}
\State $\gP \gets \text{UpdateRelativePoseFn}(\gP; \gB_+ \cup \gB_-)$
\EndWhile
\State \textbf{return} traversability function $\gT$, relative pose function $\gP$
\end{algorithmic}
\end{algorithm}

\subsection{Learning Dynamical Distances}

We aim to learn a traversability function $\gT(o_i, o_j) \in \mathbbm{R}^+$ that reflects whether \emph{any} controller can successfully navigate between observations $o_i$ and $o_j$. More precisely, we will learn to predict the estimated number of time steps required by a controller to navigate from one observation to another. This function must encapsulate knowledge of physics beyond just geometry. For example, tall grass and bushes might appear visually similar, but grass is compliant and traversable whereas bushes are not.
We explored two methods for learning this traversability function: (1) supervised learning and (2) temporal difference learning~\cite{Sutton1988, kaelbling1993learning}. To learn the distance function via supervised learning, we create a dataset $\sD_+$ of observation pairs $(o_i, o_j)$ taken from the same trajectory and regress to the number of timesteps $d_{ij} = j - i$ elapsed between these observations. The distance predicted by this approach corresponds to the estimated number of time steps required by the behavior policy (that which collected the experience) when navigating between two observations. Thus, this approach is simple but may overestimate the true shortest path distances.

The second approach to learning the distance function is via temporal difference learning~\citep{kaelbling1993learning}. This approach uses the same experience as before.
While this approach adds additional complexity, in theory it converges to the shortest path distance.
In our experiments, we found little difference between these two approaches (see Table~\ref{tab:ablations}),
but expect that the temporal difference learning approach would be important when moving to settings where the shortest path distance is much shorter than a random walk distance.

\subsubsection{Negative Mining (Key Idea 1)}
\label{sec:negatives}
In our experiments, we found that training the distance function using only observation pairs from the same trajectory performed poorly. We hypothesize that the root cause was distribution shift: when building the topological graph we must evaluate the distance function on observation pairs collected from different trajectories, possible from different times of day. To mitigate this problem, we augment the dataset by adding a new dataset $\sD_-$ obtained by sampling observations from different trajectories, labeled as $d_\text{max}$.
We find this augmentation, hereby referred to as \emph{negative sampling}, to be critical in the successful training and evaluation of $\gT$ in our experiments, offering significant improvements over prior methods.

\subsection{The Topological Graph}

We build a topological graph $\gM$ using the learned distance function together with a collection of previously-observed observations $\{o_t\}$. Each \emph{node} in the graph corresponds to one of these observations. We add weighted \emph{edges} between every node, using weights predicted by the distance function $\gT$.

\subsubsection{Graph Pruning (Key Idea 2)}
As the robot gathers more experience, maintaining a dense graph of traversability across all observation nodes becomes redundant and infeasible, as the graph size grows quadratically.
For our experiments,
we sparsify trajectories by thresholding the edges that get added to the graph: edges that are easily traversable $\left(\gT(o_i, o_j) < \delta_\text{sparsify}\right)$ are not added to the graph, since the controller can traverse those edges with high probability.

\subsubsection{Planning with the Graph}
\label{sec:pruning}
We localize the current observation $o_t$ and goal observation $o_G$ in the graph, adding direct edges (weighed by their traversability) to their corresponding ``most-traversable" neighbors.
We use the weighted Dijkstra algorithm to compute the shortest path to goal, and the immediate next node in the planned path is then handed over to the controller.

\begin{algorithm}[t]
\caption{Deploying \sysName}\label{alg:deploying}
\begin{algorithmic}[1]
\State  \textbf{Input} current image $o_t$, goal image $o_G$, and topological graph $\gM$.
\State Add $o_t, o_G$ to the map $\gM$ using distances from $\gT$.
\State $o_{w_1}, o_{w_2}, \cdots \gets \text{Dijkstra}(\text{start}=o_t, \text{goal}=o_G, \gM)$
\State Estimate relative pose of first waypoint: $\Delta p \gets \gP(o_t, o_{w_1})$
\State $u_t \gets \textsc{PD-Controller}(\Delta p)$
\State \textbf{return} control $u_t$
\end{algorithmic}
\end{algorithm}

\subsection{Designing the Controller}
\label{sec:controller}
After the planner predicts a waypoint observation, the controller must output an action that takes the agent towards that waypoint. The main challenge in navigating to this waypoint is that both the current state and waypoint are represented as high-dimensional observations (e.g., images). To address this challenge, we learn a relative position predictor $\gP$
that takes as input two observations and predicts the relative pose between these observations. We learn this relative pose predictor via supervised learning: for pairs of observations $(o_i, o_j)$ that occur nearby within the collected trajectories, we estimate the relative pose $\Delta p_{ij}$ using onboard odometry and use this relative pose as the label for learning.

The complete controller works as follows. Given the current observation and waypoint observation, we use the relative pose predictor to estimate the relative pose of the waypoint relative to the robot's current position. The robot then uses odometry and a simple PD controller to steer toward this waypoint.
We compare against alternative controllers in Section~\ref{sec:ablations}.

\subsection{Implementation Details}

Inputs to the traversability function $\gT$ and relative pose predictor $\gP$ are pairs of observations of the environment, represented by a stack of two consecutive RGB images obtained from the onboard camera at a resolution of $160\times120$ pixels. $\gT$ comprises a MobileNet encoder~\cite{howard2017mobilenets} followed by three densely connected layers to project the $1024-$dimensional latents to $50$ class labels. $\gP$ has a similar architecture as $\gT$, comprising of a MobileNet encoder followed by three densely connected layers projecting the $1024-$dimensional latents to $3$ outputs for waypoints: $\{\Delta x, \Delta y\}$. Both $\gT$ and $\gP$ use the same encoder.

We train the traversability function on $\sD_+ \cup\, \sD_-$, discretizing the timesteps $d_{ij}$ into bins $\{1, \cdots, d_\text{max} = 50\}$ and minimizing the cross entropy loss.
The relative pose predictor $\gP$ is trained on $\sD_+$ to minimize the $\ell_2$ regression loss.
We use a batch size of 128 and perform gradient updates using the Adam optimizer~\cite{Kingma2015AdamAM} with learning rate $\lambda=10^{-4}$. Algorithms~\ref{alg:training} and \ref{alg:deploying} summarize our approach in the training and deployment stages, respectively.

\section{Experiments}
\label{sec:experiemnts}

We designed our experiments to answer three questions:
\begin{enumerate}[label={\bf Q\arabic{*}.}, leftmargin=2.8\parindent]
    \item How does \sysName compare to prior methods for the task of goal-conditioned visual navigation from offline data?
    \item Does \sysName generalize to novel environments? Can it adapt on the fly?
    \item What are the alternate design choices for the controller and how do they compare against our choice in Section~\ref{sec:controller}?
\end{enumerate}

\begin{figure}
    \centering
    \includegraphics[width=0.8\columnwidth]{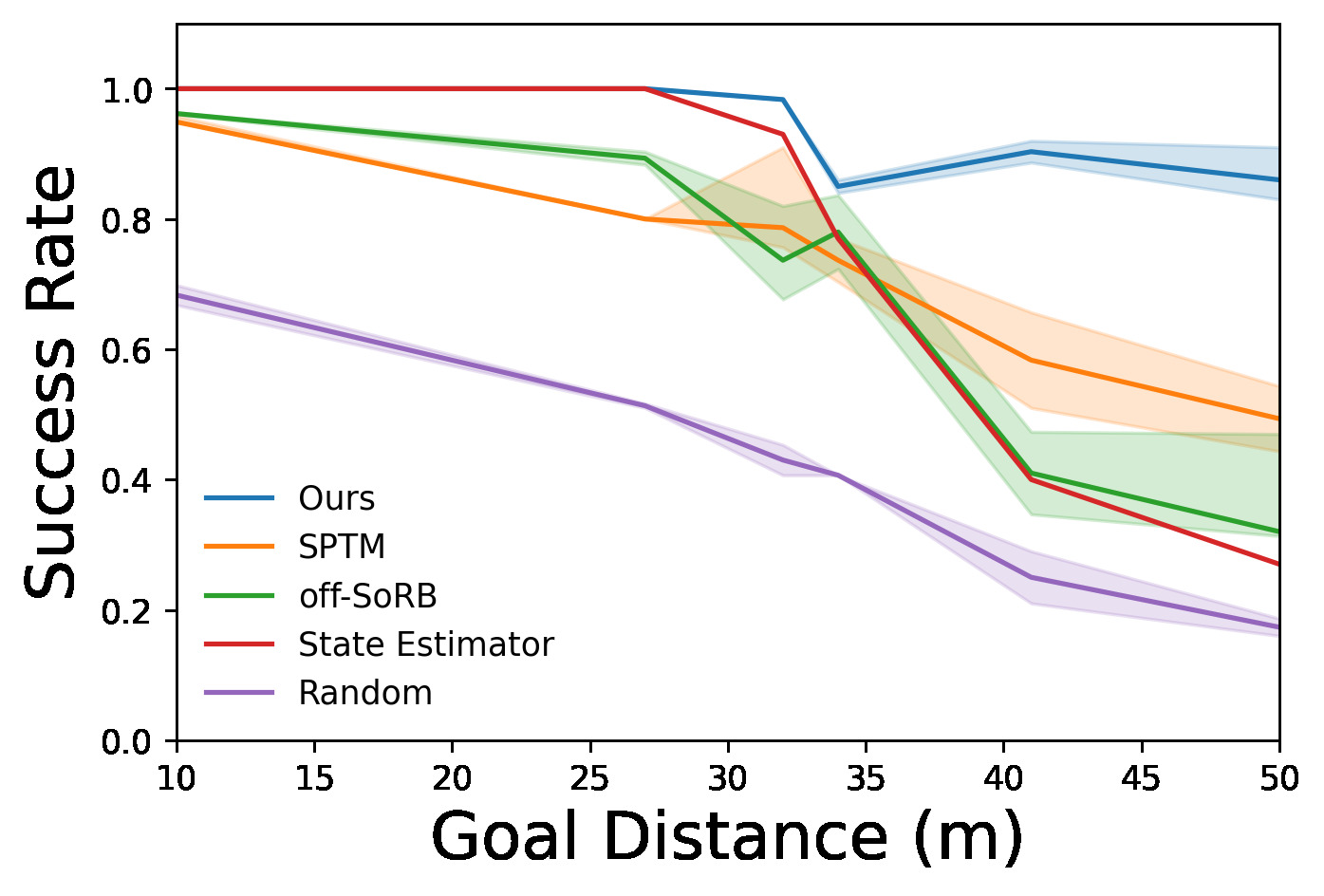}
    \caption{\textbf{Real-World Navigation}: While all non-random methods successfully reach nearby goals, only \sysName reaches goals over 40 meters away. Here, success rate is defined as the average over portion of the expert trajectory to goal that each run successfully completes.}
    \label{fig:real_success}
    \vspace*{-1.5em}
\end{figure}

\subsection{Goal-Conditioned Visual Navigation from Offline Data}
\label{sec:core_results}
We perform our evaluation in a real-world outdoor environment consisting of urban and off-road terrain. We train on $40$ hours of data that was gathered in prior work~\cite{kahn2020badgr} over 10 months prior to the experiments in this paper. The data shows significant variation in appearance due to seasonal changes (see Fig.~\ref{fig:realworld}); learning navigational affordances and traversability would require the algorithms to discard the irrelevant modes of variance (e.g., appearance) and establish correspondence across seasons and times of day.

Since this evaluation takes place in the real world, we do not have the luxury of training online RL policies or transfer from simulation. We evaluate \sysName against four baselines:
\begin{enumerate}
    \item[] {\em SPTM}: a dense topological graph combined with a controller that maps observation pairs to motor commands, trained via supervised learning~\cite{savinov2018sptm}
    \item[] {\em off-SoRB}: an offline variant of SoRB that uses a  topological graph and offline RL to learn a distributional Q-function~\cite{eysenbach2019sorb}
    \item[] {\em State Estimator}: a na\"ive baseline that uses a state estimator network that regresses observations to ground-truth state $(x, y, \theta)$, followed by a position controller; note that this baseline has access to true position (from GPS), which is not available to our method
    \item[] {\em Random}: a random walk, as described in Section~\ref{sec:controller}
\end{enumerate}
While there have been other successful instantiations of methods combining planning and learning, they make some limiting assumptions that make them difficult to apply to our problem setting. \emph{LSTN}~\cite{meng2020scaling} uses a photorealistic simulator to train its distance and action models, using $\sim1.5$M samples, while \emph{PRM-RL}~\cite{francis2020longrange} uses a 3D kinematic simulator
simulation replica to train a reactive controller, coupled with physical rollouts in the real world to build a PRM. \sysName does not assume access to any simulator, and learns directly from offline real data.

Towards answering {\bf Q1}, we evaluate the goal-reaching performance of \sysNamenosp. We select 6 \{start, goal\} image pairs in the original urban environment and compare the goal reaching performance of each method (avg. of 3 trials).
We report the success metric as the average over portion of the expert trajectory to goal that each run successfully completes. %

\begin{figure}
    \centering
    \includegraphics[width=0.9\columnwidth]{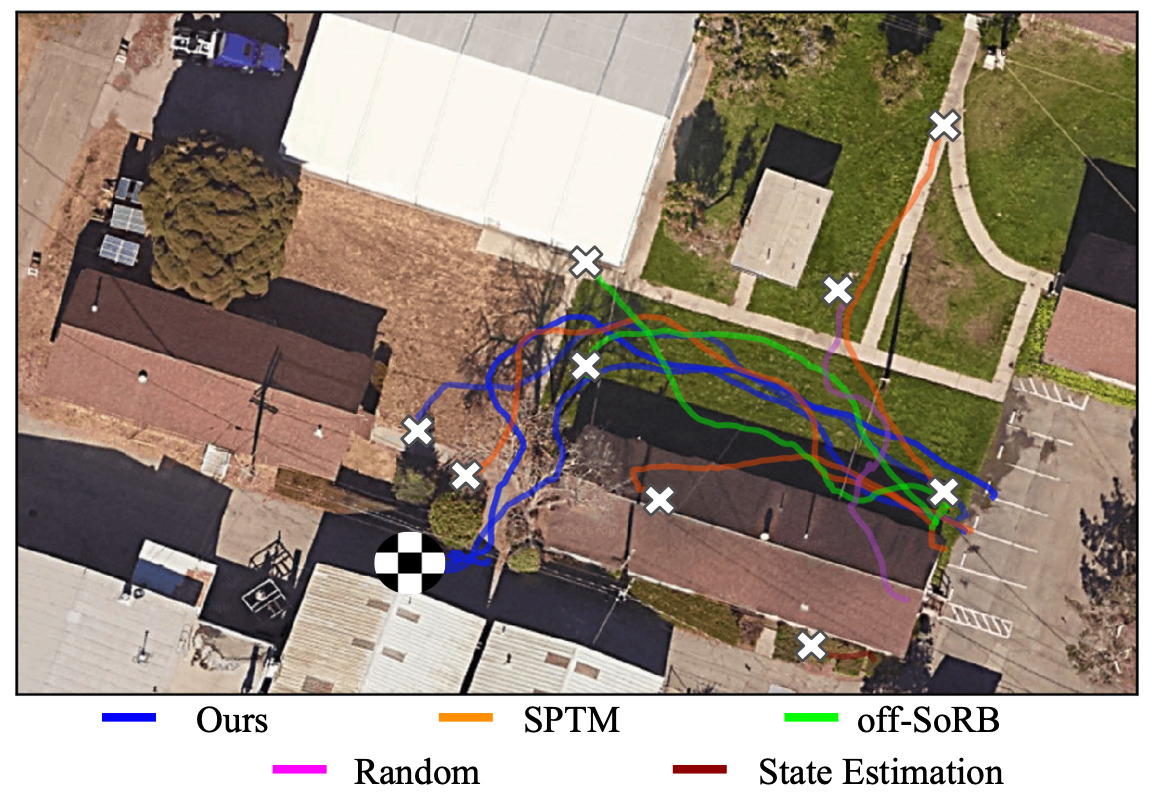}
    \caption{\textbf{Qualitative Results in the \texttt{urban} Environment}: Each approach was directed to a visual goal  $\sim50$m away (marked by checkerboard circle) -- with 3 runs per approach. \sysName is the only approach that is consistently able to reach the goal while avoiding collisions or getting stuck. }
    \label{fig:real_qualitative}
    \vspace*{-1.5em}
\end{figure}

As shown in Fig.~\ref{fig:real_success}, \sysName performs well on all tasks, achieving a success rate of 86\% on even the most challenging tasks.
As expected, the random baseline, which ignores the goal, fails to reach most goals. The state estimator baseline performs a bit better, but struggles to reach more distant goals because it is not reactive, and hence cannot take actions to avoid collisions. Off-SoRB performs well on nearby goals, but as the goals get increasingly difficult to reach, it is unable to follow the planned trajectory. Visualizing the topological graph built by SoRB uncovers many disconnected components,
resulting in no path to goal. We hypothesize that this is attributed to the difficulty in training Q-functions from offline data. SPTM, which uses supervised learning instead of Q-learning,
is effective at solving the task on shorter horizons and outperforms off-SoRB on longer horizons. However, \sysName still performs substantially better on all goal distances, especially those over 30 meters. We attribute these improvements to the additional negative sampling and graph pruning techniques discussed in Section~\ref{sec:method}.
We visualize trajectories in Fig.~\ref{fig:real_qualitative}.

\subsection{Generalization and Adaptation}

\begin{figure}[ht]
    \centering  
    \includegraphics[width=\columnwidth]{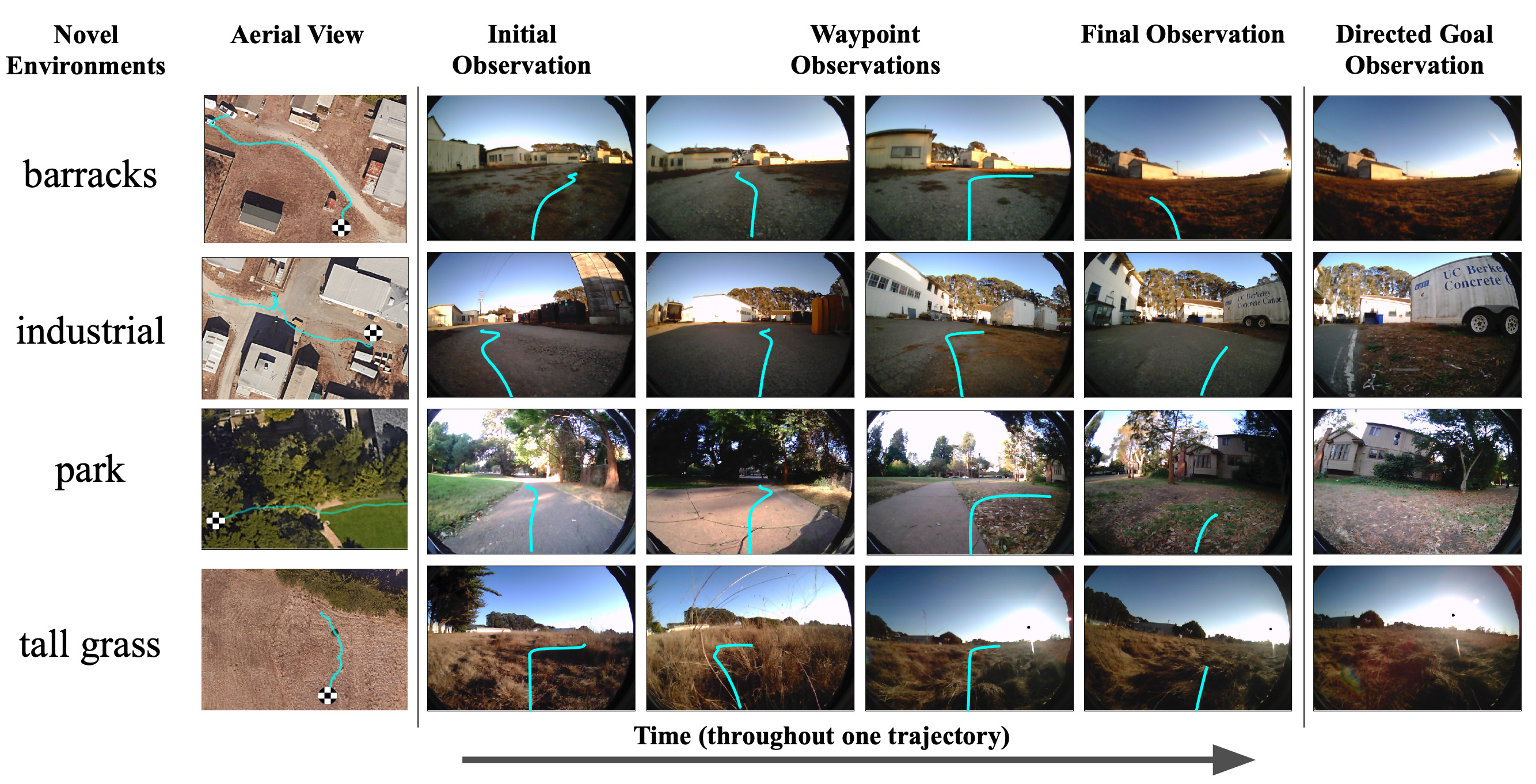}
    \caption{\textbf{Generalization Experiments}: We evaluate \sysName in four new outdoor environments. For each, we collect a few dozen minutes of experience to adapt the distance function and relative pose predictor. Then, given a goal image (last column, checkerboard location in aerial view), the robot attempts to navigate to the goal. Columns $4 - 7$ indicate that the robot succeeds in reaching the goal image. Cyan lines indicate the actions taken by \sysNamenosp.}
    \label{fig:generalization_qual}
\end{figure}

The experiments in the previous section evaluate navigation to new goals in a previously seen environment. In this section, we additionally evaluate how quickly \sysName can adapt to an entirely new, unseen environment, by constructing a new graph and finetuning the models. We use the four settings shown in Fig.~\ref{fig:generalization_qual}, all of which are distinct from the setting used in our main experiments (Sec.~\ref{sec:core_results}). In each new environment, a human controlled the robot to provide initial exploration data.
After this initial data collection, the robot collected experience \emph{autonomously}: it randomly sampled a previously-observed image as  thegoal and used \sysName to attempt to reach this goal. After each episode, we used all experience from the new environment (both the expert trajectories and the self-collected trajectories) to finetune $\gT$ and $\gP$. We refer to this approach to generalization as \sysName-Finetune.

\begin{figure}
    \centering
    \includegraphics[width=0.8\columnwidth]{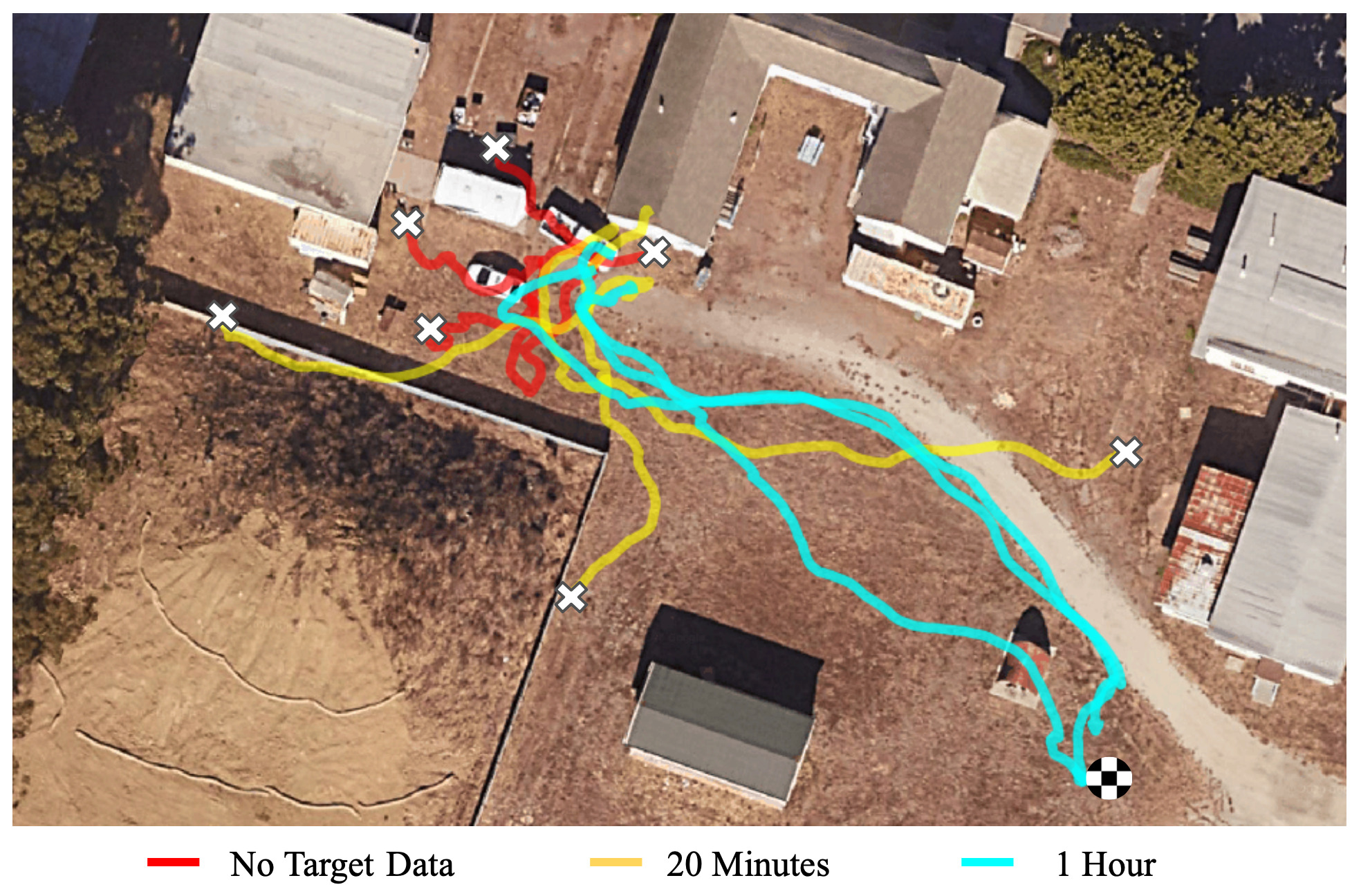}
    \caption{\textbf{Fast Adaptation to a New Environment}:
    After training \sysName in one environment, we deploy the system in a novel environment, shown above. By practicing to reach self-proposed goals and using that experience to finetune the controller, \sysName is able to quickly gain competence at reaching distance goals in this new environment, using just 60 minutes of experience. Example rollouts towards a goal $35$m away (marked by checkerboard circle) demonstrate \sysName self-improving from interactions in the \texttt{barracks} environment.}
    \label{fig:gen_qualitative}
\end{figure}

In Fig.~\ref{fig:generalization_qual} we visualize trajectories after 60 min of data collection in the new environment and observe that the robot successfully reaches the goal in most cases. We emphasize that these environments are considerably different from those used in Sec.~\ref{sec:core_results}, on which our models were initially trained. To illustrate the learning dynamics in this generalization setting, we plot self-collected rollouts after 0 minutes, 20 minutes, and 60 minutes of practice in the new environments. As shown in Fig.~\ref{fig:gen_qualitative}, the robot's performance in the new domain gets progressively better with more (autonomous) practice; after 60 minutes it succeeds in reaching the goal in all three attempts.

Table~\ref{tab:gen_table} summarizes the success rate on the generalization task of our method and two alternative versions of \sysNamenosp. \sysNamenosp-Source directly uses the traversability function and relative pose function trained in the source domain (Sec.~\ref{sec:core_results}), without incorporating any experience from the new environment. In contrast \sysNamenosp-Target learns these same models using only experience from the new ``target'' domain, without leveraging any of the previously-collected experience. \sysNamenosp-Finetune outperforms these baselines, highlighting the importance of combining old and new experience. As an additional baseline, we take the SPTM model from Sec.~\ref{sec:core_results} and finetune it on experience from the new domain. We observe that \sysName-Finetune also generalizes better than SPTM-Finetune, We hypothesize that \sysName generalizes better than SPTM because of the additional hierarchical structure of \sysNamenosp.

\begin{table}
    \centering
     \begin{tabular}{lx{1.2cm}x{1.2cm}x{1.2cm}x{1.2cm}}
     \toprule
    {Environment} & {\sysName} & {\sysName} & {\sysName} & {SPTM} \\ 
     & {Source} & {Target} & {Finetune} & {Finetune} \\ 
    \midrule
    \texttt{barracks}         &    0.27         &        0.42     &     \textbf{0.96}   &      0.74    \\
    \texttt{industrial}    &    0.13     &      0.44      &    \textbf{0.84}    &     0.68    \\
    \texttt{park}          &    0.04     &      0.32   &   \textbf{0.82}    &   0.71   \\
    \texttt{tall grass}    &    0     &     0.38    &   \textbf{0.79}  &  0.56     \\
    \bottomrule
    \end{tabular}
    \caption{\textbf{Generalization Results}: Our approach to generalization (``\sysName-Finetune'') successfully navigates learns to navigate in four new environments (shown in Fig.~\ref{fig:generalization_qual}) using just 60 minutes of experience in the new environment. Baselines that use only experience from the source or target domains are substantially less successful. Applying our finetuning approach on top of SPTM shows some generalization, but is outperformed by \sysNamenosp-Finetune.}
    \label{tab:gen_table}
\end{table} %

\subsection{Comparisons to Online Methods}
While Section~\ref{sec:core_results} establishes that \sysName outperforms competitive offline methods for the task of goal-conditioned navigation, here we also investigate the performance of our method in comparison to popular online RL algorithms. Since the sample complexity of online RL algorithms forbids us from testing this in the real world, we a Unity-based, photorealistic outdoor navigation simulator.
We include new additional baselines in the simulated experiment:
\begin{enumerate}
    \setcounter{enumi}{4}
    \item[] {\em PPO}: a popular reactive controller for indoor visual navigation algorithms~\cite{schulman2017ppo, wijmans2020ddppo}
    \item[] {\em SoRB}: online version of the ``off-SoRB'' baseline~\cite{eysenbach2019sorb}
\end{enumerate}

\begin{figure}
    \centering
    \includegraphics[width=0.9\columnwidth]{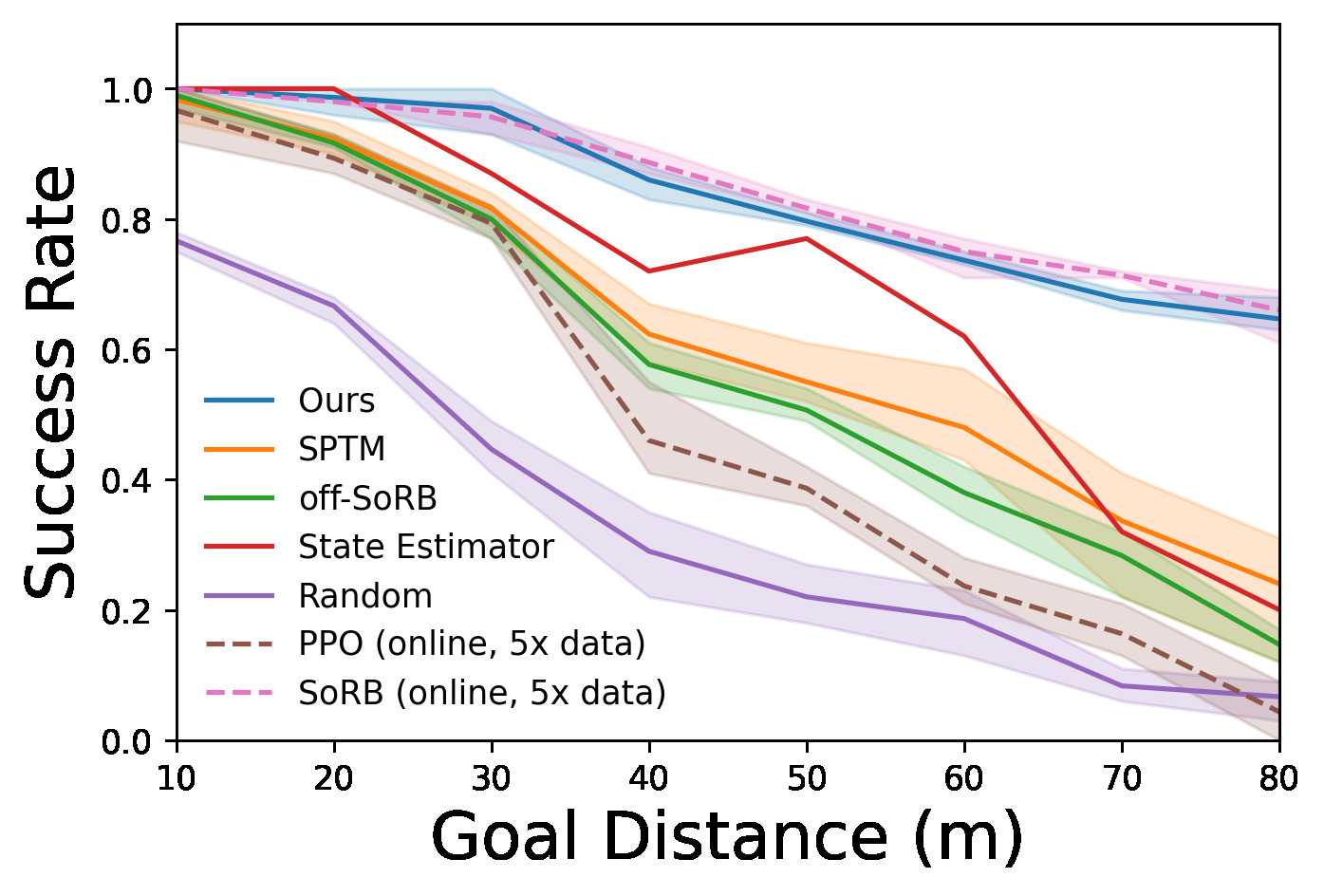}
    \caption{\textbf{Results from Simulated Navigation}: \sysName is substantially more successful at reaching distance goals than all offline baselines, while performing competitively with SoRB, a popular online baseline combining Q-learning and topological graphs. We emphasize that SoRB and PPO require $5\times$ online data collection, making them prohibitively expensive to apply in the real-world.}
    \label{fig:sim_success}
\end{figure}

We show results in Fig.~\ref{fig:sim_success}.
PPO performs poorly and is outperformed by \sysNamenosp, suggesting that a single image-based reactive policy is insufficient for solving long-horizon goal-reaching tasks, even when given access to 200 hours of online experience. SoRB outperforms other baselines and performs on par with \sysNamenosp.
However, whereas \sysName requires $40$ hours of offline data, SoRB requires $200$ hours of online data, and must recollect this data for every experiment.

\subsection{Ablation Experiments}
\label{sec:ablations}
A key design decision for \sysName that differentiates it from prior methods (e.g.,~\cite{meng2020scaling, eysenbach2019sorb}) is how the controller generates actions to reach the next waypoint.
We evaluate variants of \sysName that use alternative controllers and present results in Table~\ref{tab:ablations}.
Two simple baselines, ``direct actions'' and ``direct actions (discrete)'', use the goal-conditioned behavior cloning method of~\cite{savinov2018sptm, Ding2019} to directly predict (discrete) actions from the current and goal observations, without utilizing the topological map.
Recall that our method uses the planner to command waypoints and then uses the relative pose together with a PD controller to reach each waypoint. We compared against a baseline that uses a different low-level controllers to reach these same waypoints: ``Waypoint, Discrete'' takes actions using the ``direction actions (discrete)'' controller described above.
As an alternative training scheme, ``TD Waypoint'' is a variant of our method that learns the traversability function via TD learning instead of supervised learning. Finally, we compare to two ablations of our method that skip the graph pruning and negative sampling stages of \sysNamenosp.

\begin{table}[]
\centering
\begin{tabular}{l@{\hspace{.25cm}}c@{\hspace{0.25cm}}c@{\hspace{0.25cm}}c@{\hspace{0.25cm}}c@{\hspace{0.25cm}}c}
\toprule
& \multicolumn{5}{c}{Success Rate @ Distance $d$ (m)}\\ 
Controller & {$d\!=\!10$} & {$d\!=\!20$} & {$d\!=\!30$} & {$d\!=\!40$} & {$d\!=\!50$} \\ 
\midrule
{Direct Actions (Discrete)}   &     {0.87}        &       0.81      &      0.74       &       0.65      &       0.45      \\
{Direct Actions}   &     {0.98}        &       0.89      &      0.74       &       0.73      &       0.4      \\
{Waypoint, Discrete} &       \textbf{1.0}      &        0.95     &    0.91         &      0.82       &      0.7       \\ 
Waypoint      &      \textbf{1.0}   &     \textbf{1.0}        &     \textbf{0.95}        &       {0.88}     &      {0.81}       \\
TD Waypoint       &      \textbf{1.0}   &     \textbf{1.0}        &     \textbf{0.96}        &       \textbf{0.87}     &      \textbf{0.87}       \\ 
Waypoint, No Pruning &       \textbf{1.0}      &        \textbf{0.88}     &    0.81         &      0.79       &      0.52       \\ 
Waypoint, Only Positives &      \textbf{1.0}      &      0.91       &     0.75        &        0.76     &        0.43     \\
\bottomrule
\end{tabular}
\caption{\textbf{Ablation Experiments}: We investigate design choices for the parametrization of the controller. Using waypoints as a mid-level action space is key to the performance of \sysNamenosp, which is particularly emphasized for distant goals. While training the models, we show that \sysName can be trained with either supervised or TD learning and report similar performance. We also show that the two key ideas presented -- graph pruning and negative sampling -- are indeed essential for the performance of \sysName in the real-world.}
\label{tab:ablations}
\end{table}

\subsection{Applications and Qualitative Results}

\sysNamenosp's ability to navigate using perception and landmarks, without access to maps or localization, can enable a number of intuitive applications, which we illustrate through qualitative results in this section. We constructed two demonstrations that reflect potential applications of our system:
\begin{enumerate}
    \item \textit{Contactless Last-Mile Delivery}: We demonstrate last-mile delivery in a residential complex by using \sysName to autonomously deliver mail and food to visually-indicated delivery locations. In this setting, users specify delivery destinations for the robot simply by taking a photograph of the desired destination, and the robot autonomously navigates to this destination to deliver a package.
    \item \textit{Autonomous Inspection}:
    Densely constructed building complexes, like university campuses, are often unmapped or lack accurate spatial localization.
    We reprogram \sysName to periodically navigate to landmarks, specified as images, around the campus to set up an autonomous patrolling system. Discrepancies can be identified by comparing the observations to previous observations (stored in the topological graph).
\end{enumerate}

\begin{figure}
    \centering
    \includegraphics[width=\columnwidth]{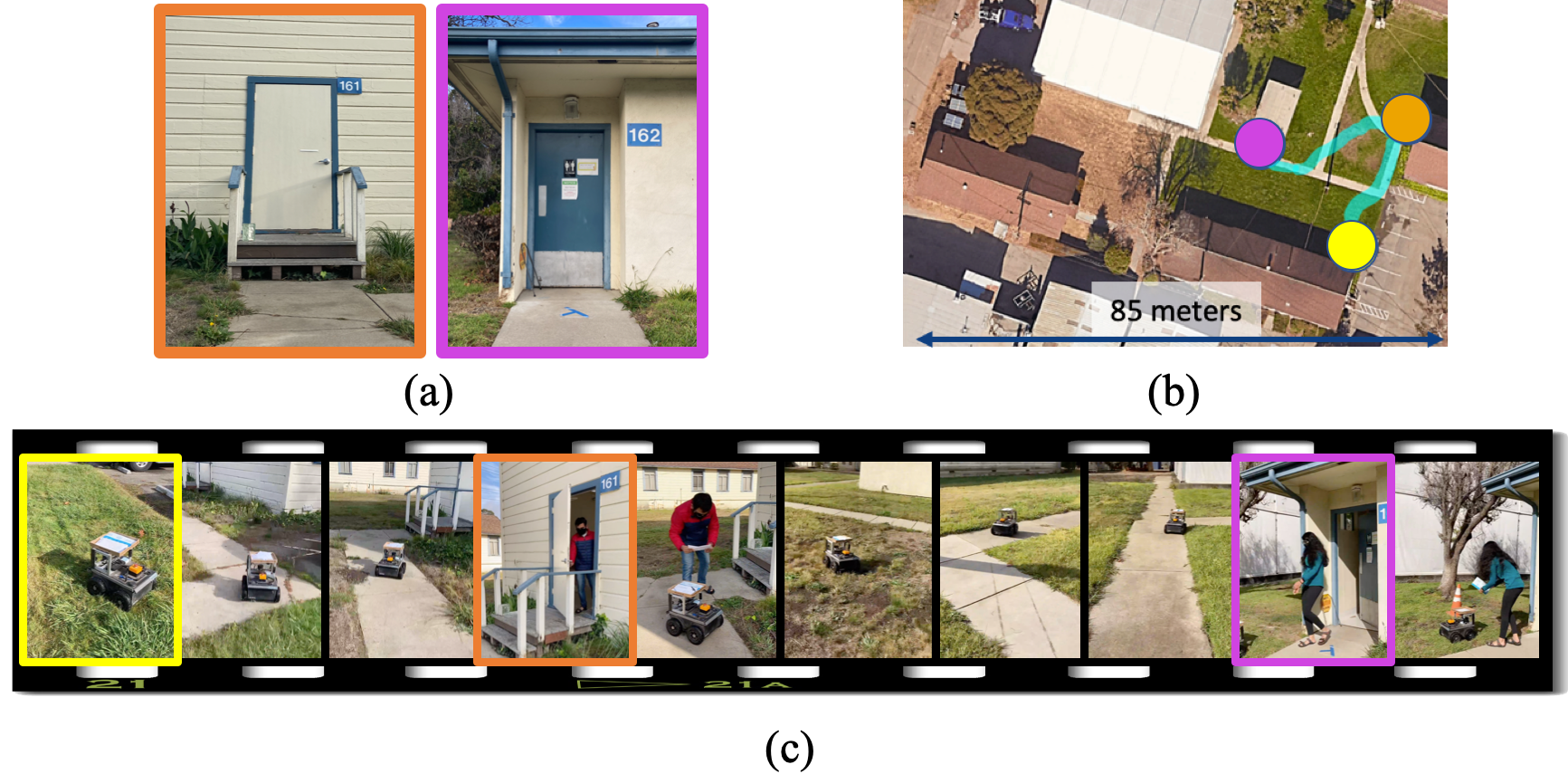}
    \caption{\textbf{Contactless Last-Mile Delivery Demo}:
    Given a set of visually-indicated goals (a), \sysName can perform contactless delivery in the \texttt{urban} neighborhood successfully, as shown in the filmstrip (c). An overhead view (b) with starting position marked in yellow and respective goals marked in orange and magenta shows the trajectory of the robot (cyan). \textit{Note: The satellite view (b) is solely for visualization and is not available to the robot.}}
    \label{fig:mailman_demo}
\end{figure}

\begin{figure}
    \centering
    \includegraphics[width=0.8\columnwidth]{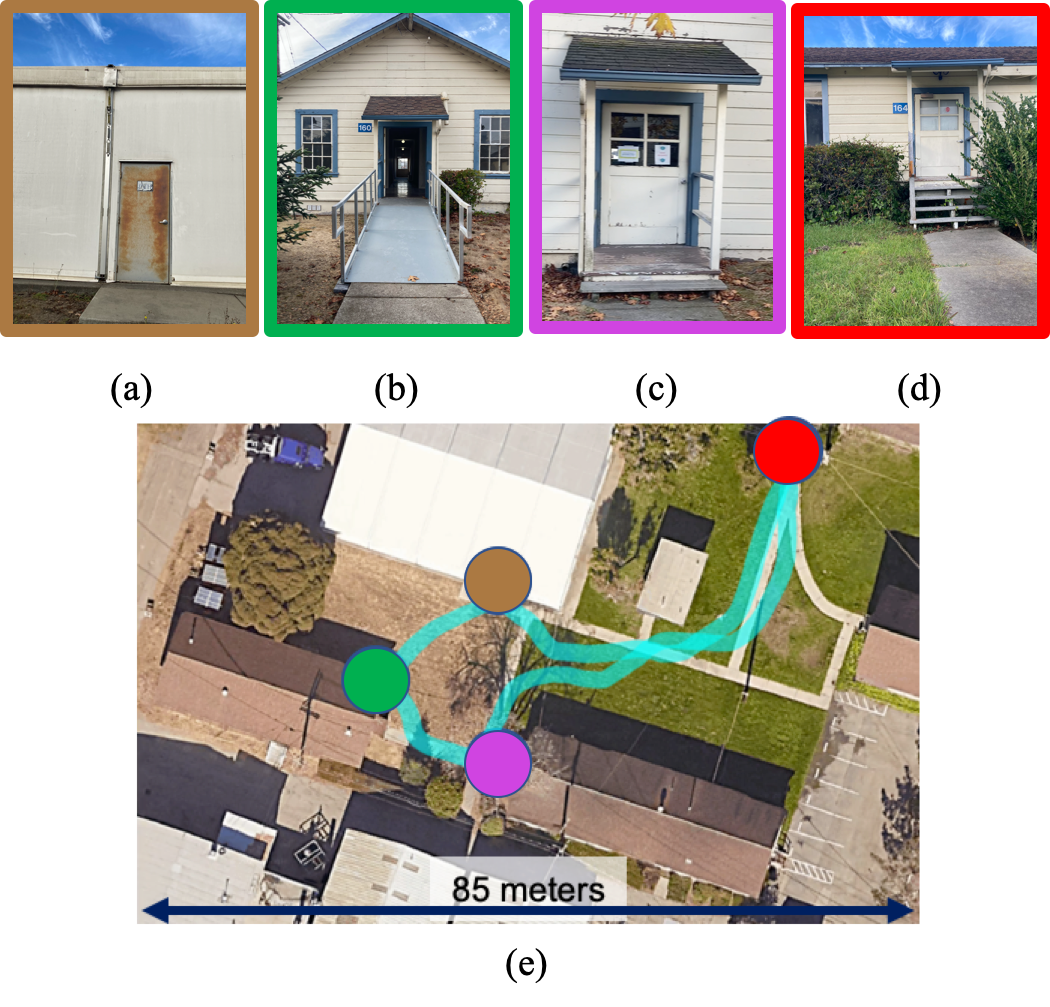}
    \caption{\textbf{Autonomous Inspection Demo}:
    Given a set of visual landmarks (a--d) in a university campus, \sysName can perform autonomous inspection by navigating to these goals periodically. An overhead view (b) shows color-coded goals and the trajectory taken by robot (cyan) in one cycle. \textit{Note: The satellite view (e) is solely for visualization and is not available to the robot.}}
    \label{fig:patrol_demo}
    \vspace*{-1em}
\end{figure}

Figures \ref{fig:mailman_demo} and \ref{fig:patrol_demo} show \sysName successfully performing these tasks in the \texttt{urban} environment. Videos of the qualitative results, generalization experiments, and real-world applications can be found at the project website (\texttt{\href{https://sites.google.com/view/ving-robot}{sites.google.com/view/ving-robot}}).

\section{Conclusion}
\label{sec:conclusion}
In this paper, we proposed \sysNamenosp: a system for goal-directed navigation using visual observations and goals on an outdoor ground robot. While conceptually similar to prior methods, we demonstrate that a few key design choices, such as pruning the topological graph, parametrizing the controller in terms of a relative pose predictor and sampling negatives while training to minimize distribution shift, allow \sysName to learn to successfully navigate using only offline experience, a setting in which many prior methods fail. Intriguingly, we also demonstrate that \sysName can be quickly adapted to navigate in new environments. These generalization and self-improvement attributes highlight that learning-based approaches are not only an effective mechanism for handling high-dimensional observations, but are also amenable to fast adaptation to novel environments. Further, we have demonstrated \sysName on a number of real-world applications in dense, urban environments that may be unmapped or GPS-denied, and specifying visual goals is convenient --  contactless last-mile delivery and autonomous inspection.

Our method requires a static, offline dataset of observations over which we can plan. Many real-world tasks are non-stationary, with the distribution of observations shifting over time (e.g., lighting changes, dynamic objects, etc.). In future work, we aim to incorporate representations of observations and goals that are robust to such distributional shifts, which would expand the generalization capabilities of our method.

\section*{Acknowledgments}
This research was funded by the Office of Naval Research, DARPA Assured Autonomy, and ARL DCIST CRA W911NF-17-2-0181, with computing support from Google and Amazon Web Services. The authors would like to thank Jonathan Fink and Ethan Stump for their help setting up the simulation environment used for developing this research. 
\renewcommand*{\bibfont}{\small}
{\bibliographystyle{IEEEtran}
\bibliography{references}}
\end{document}